\documentclass[conference]{IEEEtran}

\usepackage{cite}
\usepackage{amsmath,amssymb,amsfonts}
\usepackage{graphicx}
\usepackage{booktabs}
\usepackage{multirow}
\usepackage{url}
\usepackage{xcolor}
\usepackage{tikz}
\usetikzlibrary{arrows.meta,positioning,fit,backgrounds,calc}
\usepackage{pgfplots}
\pgfplotsset{compat=1.17}
\usepackage{enumitem}
\usepackage{algorithm}
\usepackage{algpseudocode}

\newcommand{\genmodel}{Gemma\,4:31B}
\newcommand{\dsrc}{\mathcal{D}_{\mathrm{src}}}
\newcommand{\dtgt}{\mathcal{D}_{\mathrm{tgt}}}
\newcommand{\ctgt}{\mathcal{C}_{\mathrm{tgt}}}

\newcommand{\gr}[1]{\textcolor{violet!60}{#1}}

\begin{document}


\title{No Audio, No Transcripts, No Problem: LLM-based ASR Adaptation Using Only Domain Documents}


\author{
\IEEEauthorblockN{
Sergio Burdisso$^{1}$,
Dairazalia Sánchez-Cortés$^{1}$,
Esaú Villatoro-Tello$^{1}$,
Thibault Bañeras-Roux$^{1}$, \\
Manjunath K E$^{2}$,
Kadri Hacioglu$^{2}$,
Petr Motlicek$^{1,3}$,
Andreas Stolcke$^{2}$
}
\IEEEauthorblockA{
$^{1}$\textit{Idiap Research Institute, Martigny, Switzerland} 
$^{2}$\textit{Uniphore, U.S.A.}  
$^{3}$\textit{Brno University of Technology, Brno, Czech Republic}
}
}

\maketitle

\begin{abstract}
Can a speech recognizer be adapted to a new domain with no target audio and no transcripts, using only written documentation?
This is the realistic onboarding setting: a new customer hands over raw documentation, not the transcripts that text-only adaptation normally assumes.
We turn a domain's documents into adaptation text in two ways, directly as written sentences or rendered by a large language model (LLM) into synthetic spoken dialogs, and feed them to a recently proposed denoising-based text-only adaptation method.
We show that, across in-domain, out-of-domain, and cross-domain settings on two corpora, document-only adaptation can match or closely approach real transcriptions; whether written or conversational rendering is best depends on speaking style and the amount of text generated.
To our knowledge, this is the first study of document-only adaptation of LLM-based ASR. As no existing corpus pairs audio, transcripts, and documents, we also built and released a document-grounded benchmark.
\end{abstract}


\section{Introduction}
\label{sec:intro}
LLM-based ASR systems, which connect a pretrained speech encoder to a frozen large language model (LLM) through a lightweight learnable projector, have become a powerful approach to transcription~\cite{ma2024embarrassingly, kumar2024performance, 10445874, 11434631, prompt_projection_2026,villatoro_2026_context_proj}. 
However, these systems still degrade under domain shift and transcribing target-domain audio for model adaptation is expensive.
\emph{Text-only adaptation}, which specializes the system using text from the target domain, is therefore an appealing alternative given the abundance of text relative to transcribed speech~\cite{10389722}.

A recurring and largely unexamined assumption underlies this line of work: the available target-domain text consists of \emph{ground-truth transcriptions} of spoken conversations.
Prior text-only adaptation methods for conventional end-to-end ASR~\cite{sato22_interspeech, bataev23_interspeech, kurian2025domain} and for LLM-based ASR~\cite{fang2025low, ma2024effective, burdisso2025denoising, banerasroux_2026, burdisso2025denoising-interspeech} all adapt on transcripts.
Yet in the most realistic onboarding scenario (a new client hands over their materials so an ASR can be delivered for their domain), transcripts of in-domain calls are exactly what is \emph{not} available.
What is available is \emph{raw documentation}: company websites, product and service descriptions, manuals, policies, etc.
Obtaining transcripts instead presupposes the very audio collection and annotation that text-only adaptation is meant to avoid.
Beyond practicality, document-based adaptation suits privacy-aware deployment: adapting from a company's existing documentation avoids recording and transcribing sensitive customer conversations.

All this motivates the central question of this paper: \emph{can an ASR system be adapted to a new domain using only raw documents, with no target-domain audio and no target-domain transcripts?}
We answer this question in the affirmative.
Our key insight is that documents and transcripts differ not only in content but in \emph{form}: a document states facts in written prose, whereas a transcript realizes those facts as spontaneous, often disfluent, turn-based speech.
We therefore turn the domain's documentation into adaptation text in two ways: used directly as written sentences or rendered by an LLM into synthetic spoken dialogs (a conversational grounded in the document), and feed them to the multi-view denoising adaptation method of~\cite{burdisso2025denoising}.
We find that document-only adaptation is competitive with ground-truth transcriptions across domains and regimes, and that whether written or conversational rendering is preferable depends on the target's speaking style (\emph{form}) and the amount of generated text (\emph{quantity}).
To expose the \emph{form} dependence, we evaluate on two deliberately contrasting corpus types, spontaneous customer--agent conversations and presentation-style talks and lectures: their registers sit at opposite ends of the speaking-style spectrum, with the conversational corpus matching the synthetic dialogs and the presentation corpus being closer in style to the original documents.
Our contributions are:
(a) We open up a \emph{new regime for text-only adaptation}, framed by the question above: adapting from a domain's written documentation alone. This is the most challenging case, as well as the most realistic, based on what a new client can provide; to our knowledge this scenario has not been studied before;
(b) We show this \emph{regime is viable}: rendering documents into adaptation text, as their own written sentences or as zero-shot LLM-generated dialogs, recovers most or all of the real-transcription gain across in-, out-of-, and cross-domain transfer, using two distinct speech genres;
(c) We \emph{characterize when each rendering is best} across the different settings, showing the choice turns not on speaking style (\emph{form}) alone, but also on the amount of adaptation text (\emph{quantity});
(d) We release a \emph{document-grounded benchmark}: as no public corpus pairs audio, transcripts, and documents, we reconstruct the document side from transcripts under a coverage constraint and release aligned audio--transcript--document triples for SlideSpeech~\cite{wang2024slidespeech}, along with the code.\footnote{\url{https://github.com/idiap/llm-asr-document-only-adaptation}}

\section{Related Work}
\label{sec:related}
\textbf{Text-only adaptation of ASR.} For conventional end-to-end ASR, text-only adaptation has been approached via intermediate-CTC adaptation~\cite{sato22_interspeech}, external language-model fusion~\cite{sriram2017coldfusiontrainingseq2seq, meng2021minimum,hori2025delayed}, text-to-spectrogram generation~\cite{bataev23_interspeech}, and decoder adaptation for Whisper~\cite{kurian2025domain}.
For LLM-based ASR, fine-tuning the LLM on text alone disrupts the speech--text alignment learned by the projector, causing catastrophic forgetting~\cite{fang2025low, ma2024effective, burdisso2025denoising}.
Fang et al.~\cite{fang2025low} mitigate this by monitoring alignment and stopping before collapse, while Ma et al.~\cite{ma2024effective} learn soft-prompt embeddings to replace the speech input.
The multi-view denoising approach of~\cite{burdisso2025denoising}, our adaptation backbone, preserves alignment by mixing source audio with synthetically noised text per batch.
Crucially, all of these methods adapt on \emph{ground-truth transcriptions}; none consider raw documents as the only available signal.

\textbf{Synthetic conversational data generation.} A growing body of work generates synthetic conversations to overcome the data scarcity and privacy constraints of real call-center corpora: eliciting natural customer-support dialogs~\cite{gung2023natcs}, converting raw recordings into structured datasets~\cite{echeverria2025call2instruct}, recognizing  entities in phone-call transcripts~\cite{mohammadi2026lingvarbench}, and synthesizing dialogs for downstream systems~\cite{suresh2025diasynth}, with further work on adaptive prompting~\cite{gao2023adaptive}, multi-document grounding~\cite{lee2024multi}, and user-diversity simulation~\cite{ahmad2025simulating,dou2026finua}.
These methods aim to build or augment \emph{dialog systems}, and their grounded variants typically rely on retrieval and passage segmentation~\cite{shuster2021retrieval,lee2024multi,oche2025systematic}.
In contrast, our purpose is \emph{ASR domain adaptation}: we generate spoken-style transcripts directly from raw documentation, with no retrieval or passage segmentation. We find that simple zero-shot generation over the whole document is highly effective for this task, as it preserves discourse coherence and avoids retrieval-stage errors.

\textbf{Synthetic data for ASR.} A long line of work reduces the audio burden in ASR by synthesizing data, e.g., text-to-speech augmentation and text-conditioned generation of acoustic representations~\cite{li2018training,rosenberg2019speech,rossenbach2020generating,laptev2020you,bataev23_interspeech}. 
Our setting differs in both \emph{objective} and \emph{constraint}: we do not augment an existing in-domain text distribution, since no target-domain spoken text exists at all, and we must \emph{induce} a spoken distribution from non-spoken documents, evaluating the result by its \emph{downstream value} for text-only ASR adaptation rather than by generation quality alone.

\section{Method: Adapting an ASR from Documents}
\label{sec:method}
\subsection{Problem setting}
Let $\dsrc=\{(a_i,t_i)\}$ be a source-domain corpus of paired audio $a_i$ and transcripts $t_i$, used to train the base LLM-based ASR.
For the target domain we are given \emph{only} a collection of raw documents $\ctgt=\{d_1,\dots,d_M\}$, with no target audio and no target transcripts.
The goal is to adapt the system to the target domain using $\dsrc$ and $\ctgt$ alone.
This is strictly harder than the standard text-only setting, which assumes a target transcript set $\dtgt=\{t_j\}$.

\begin{figure}[t]
\centering
\resizebox{\columnwidth}{!}{%
\begin{tikzpicture}[
  font=\scriptsize,
  box/.style={draw,rounded corners,align=center,inner sep=3pt,minimum height=7mm},
  data/.style={draw,align=center,inner sep=3pt,fill=black!4},
  proc/.style={box,fill=violet!8},
  >=Latex
]
\node[data] (tr) {Reference\\transcripts};
\node[proc,right=4mm of tr] (clu) {Cluster\\+ label};
\node[proc,right=4mm of clu] (ner) {Entity /\\keyword\\extraction};
\node[proc,right=4mm of ner] (docgen) {Doc.\\generation\\($\geq$90\% cov.)};
\node[data,right=4mm of docgen] (doc) {Raw\\documents};
\draw[->] (tr)--(clu); \draw[->] (clu)--(ner); \draw[->] (ner)--(docgen);
\draw[->] (docgen)--(doc);
\begin{scope}[on background layer]
\node[draw,dashed,rounded corners,fit=(tr)(clu)(ner)(docgen)(doc),
  inner sep=4pt,label={[font=\scriptsize\itshape]above:Benchmark construction (Sec.~IV) -- builds the released documents}] {};
\end{scope}
\node[proc,below=10mm of docgen] (d2d) {Doc.$\rightarrow$dialog\\LLM (\genmodel)};
\node[data,left=5mm of d2d] (syn) {Synthetic\\transcripts};
\node[proc,left=5mm of syn] (adapt) {Multi-view\\denoising\\adaptation};
\node[data,left=5mm of adapt] (asr) {Adapted\\LLM-ASR};
\draw[->] (doc)--(d2d);
\draw[->] (d2d)--(syn); \draw[->] (syn)--(adapt); \draw[->] (adapt)--(asr);
\begin{scope}[on background layer]
\node[draw,rounded corners,fit=(asr)(adapt)(syn)(d2d),
  inner sep=4pt,label={[font=\scriptsize\itshape]below:Document-only adaptation -- no target audio or transcripts}] {};
\end{scope}
\end{tikzpicture}}
\caption{Overview. \emph{Top (dashed), benchmark construction:} no corpus pairs audio, transcripts, and documents, so we synthesize the document side offline from reference transcripts and release it (Sec.~\ref{sec:benchmark}); this only builds the benchmark, and the adaptation method never sees these reference transcripts. \emph{Bottom, document-only adaptation:} the method uses \emph{no target audio or transcripts}, taking only the target's documents as input (here rendered into dialogs).}
\label{fig:pipeline}
\end{figure}
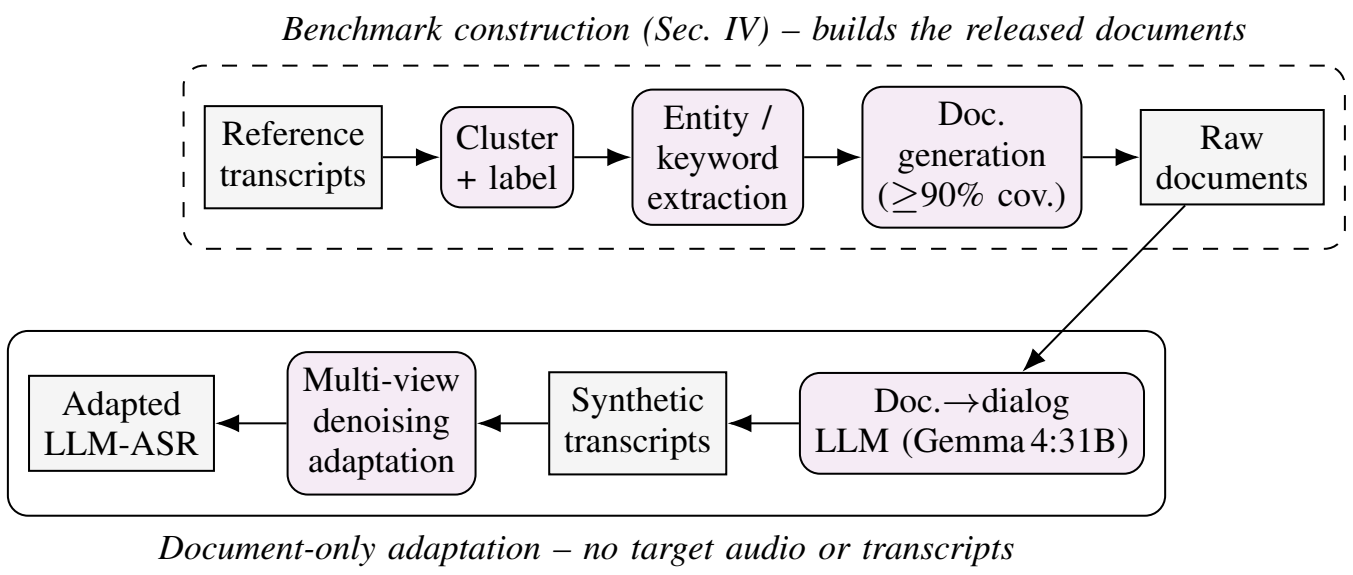

\begin{figure}[t]
\centering
\footnotesize
\fbox{\parbox{0.94\columnwidth}{
\textit{\textbf{Source document (excerpt).}} ``[\,\dots] The \gr{SilverShield
Premier plan} covers in-patient hospitalization up to \gr{\$50{,}000 per year}.
A \gr{30-day waiting period} applies before claims can be submitted. To file a
claim, customers [\,\dots]''\\[2pt]
\rule{\linewidth}{0.3pt}\\[2pt]
\textit{\textbf{Generated dialog (excerpt).}}\\
\texttt{AGENT}: THANKS FOR CALLING \gr{SILVERSHIELD} HOW CAN I HELP\\
\texttt{CUST}: HI UM YEAH I I WANTED TO ASK ABOUT THE \gr{PREMIER PLAN}\\
\texttt{AGENT}: SURE\\
\texttt{AGENT}: SO THE \gr{SILVERSHIELD PREMIER PLAN} IT COVERS HOSPITAL STAYS\\
\texttt{AGENT}: UP TO \gr{FIFTY THOUSAND DOLLARS A YEAR}\\
\texttt{CUST}: OH OKAY\\
\texttt{CUST}: AND CAN I CLAIM RIGHT AWAY\\
\texttt{AGENT}: NOT QUITE THERE'S A \gr{THIRTY DAY WAITING PERIOD} FIRST\\
\texttt{CUST}: HMM OKAY GOOD TO KNOW\\
\mbox{}\hfill\dots\hfill\mbox{}
}}
\caption{A generated dialog grounded in a source document. ``[\,\dots]'' marks
omitted document text and the trailing ``\dots'' indicates the dialog continues.
\gr{Colored} spans mark facts shared by both views, realized verbatim or in
normalized spoken form: document facts are preserved but rendered as short,
spontaneous, uppercase segments with spelled-out numbers, matching the
conversational ASR convention.}
\label{fig:example}
\end{figure}

\subsection{From documents to synthetic transcripts}
\label{subsec:doc2dialog}
We instantiate the missing transcripts $\dtgt$ by generating synthetic conversational transcripts that are faithful to the documents' content but realized as spontaneous speech.
To produce one transcript, we sample a document from the collection $\ctgt$ and prompt an instruction-tuned LLM (\genmodel) to generate a telephone call between an \textsc{Agent} and a \textsc{Customer} for a call center operating in the domain described by that document.
The generation prompt enforces four properties that make the output usable as ASR adaptation text (Fig.~\ref{fig:pipeline}):
\begin{enumerate}[leftmargin=1.2em,itemsep=0.15ex,topsep=0.3ex]
\item \textbf{Document grounding.} All factual content (product and service names, plans, policies, fees, procedures) must be drawn exclusively from the document, preferring the same terminology; The model is encouraged to avoid introducing new facts beyond those in the document.
\item \textbf{Turn-taking-like segmentation.} Output is segmented into short speech units rather than full sentences, with irregular, non-alternating speaker sequences and standalone backchannels, reflecting the turn-taking structure of conversational speech as produced by ASR segmentation.
\item \textbf{Spontaneous speaking style.} The text contains restarts, repetitions, hesitations, fillers, and other disfluencies characteristic of spontaneous speech rather than planned or read speech.
\item \textbf{ASR-style text normalization.} Numbers, dates, and email addresses are spelled out in full spoken form, initialisms such as ``USA'' and ``FBI'' are spelled out letter by letter (``U S A'', ``F B I'') while other abbreviations are expanded (e.g. "Dr." as "Doctor"), and match the transcription format of the corpora (e.g. uppercased text, no punctuation).
\end{enumerate}
For each target domain we generate $N$ synthetic dialogs.
In our experiments, to compare fairly against adaptation on ground-truth transcripts, we set $N$ to the number of ground-truth dialogs the domain has in our experimental data (Section~\ref{sec:setup}). 
Each of the $N$ generations independently samples a document uniformly at random from $\ctgt$ and produces one grounded dialog from it; consequently a single document may seed several dialogs, and the number of generated dialogs is \emph{decoupled} from the number of documents $M$ (we do not assume one dialog per document).
We match the \emph{number of dialogs} rather than the number of segments or total duration: since each generation yields a full dialog of variable length, the segment count cannot be easily controlled, so dialog count serves as a simple, comparable budget between ground-truth-transcript and synthetic adaptation.

We use zero-shot generation (no in-domain example dialogs given in the prompt), which keeps the method free of any real target transcripts; we analyze the effect of few-shot prompting in Section~\ref{subsec:nshot}.
The model is asked to return a single JSON object containing an ordered list of \texttt{\{speaker, text\}} segments (speaker $\in$ \{\textsc{Agent}, \textsc{Customer}\}), from which we use each \texttt{text} as an individual transcript; the document is supplied in Markdown so the model can exploit its headings and lists to locate key terms.
We decode with a relatively high sampling temperature ($2.0$) to encourage diverse, spontaneous output across the $N$ generations.
The resulting synthetic transcripts (\textit{i.e.} all the \texttt{text} segments) play the role of $\dtgt$ in the adaptation step.
Fig.~\ref{fig:example} illustrates the transformation: the same factual content present in a document is re-expressed as short, disfluent, spoken-style segments with numbers spelled out, a ``conversational view'' of the document.

\subsection{Adaptation backbone}
\label{subsec:backbone}
Given the synthetic target transcripts, we adapt the LLM-based ASR with the multi-view text-denoising method of~\cite{burdisso2025denoising}, which we treat as a fixed, previously validated component that is summarized below; see~\cite{burdisso2025denoising} for full details.
The method interprets LLM-based ASR as a \emph{denoising} task: the learnable projector $sp(\cdot)$, which maps the speech encoder's frame-level acoustic features for an utterance $a$ into a sequence of vectors in the LLM input-embedding space, produces a sequence of soft tokens that the frozen LLM reads as a noisy, text-like rendering of the transcript $t$ and then ``cleans'' into $t$.
Text-only adaptation therefore amounts to training the LLM to denoise corrupted transcripts $(\mathit{noise}(t),t)$, with no audio required.
Directly fine-tuning on target $(\mathit{noise}(t),t)$ pairs, however, collapses the speech--text alignment (catastrophic forgetting); the method avoids this by composing each mini-batch from four views with proportions $\sigma_a,\sigma_{ta},\sigma_t,\tau$ that sum to one:
\begin{itemize}[leftmargin=1.2em,itemsep=0.15ex,topsep=0.3ex]
\item $\sigma_a$: source audio--text pairs $(sp(a),t)$, $(a,t)\in\dsrc$;
\item $\sigma_{ta}$: projector-induced noisy source transcripts $(\mathit{noise}_a(t),t)$, where $\mathit{noise}_a(t)$ maps $sp(a)$ to its nearest LLM-vocabulary tokens (an empirical ``optimal'' noise);
\item $\sigma_t$: synthetically noised source transcripts $(\mathit{noise}(t),t)$, $t\in\dsrc$ (a cheap, audio-free approximation);
\item $\tau$: synthetically noised \emph{target} domain transcripts $(\mathit{noise}(t),t)$, which drive adaptation.
\end{itemize}
The three source views preserve alignment and avoid forgetting, while $\tau$ controls adaptation strength.
We use the same training prompt as the backbone, \texttt{Transcribe speech to text.
Speech: \{input\}}, with \texttt{\{input\}} set to $sp(a)$, $\mathit{noise}_a(t)$, or $\mathit{noise}(t)$ during training and to $sp(a)$ at inference.
We also reuse the backbone's noise function $\mathit{noise}(\cdot)$ (random character substitution then duplication; see~\cite{burdisso2025denoising} for the exact rates).
In our setting the target transcripts are the synthetic document-derived dialogs of Section~\ref{subsec:doc2dialog}.
Whereas the original method~\cite{burdisso2025denoising} sets $\tau$ per domain by a size heuristic, we fix $\tau=0.5$ for \emph{all} experiments (so $\sigma_a=\sigma_{ta}=\sigma_t=(1-\tau)/3\approx0.17$), and keep the backbone otherwise unchanged (including noise settings and all optimization/training hyperparameters), so that the \emph{text source} (ground-truth transcripts vs.\ raw document sentences vs.\ conversational renderings) is the only factor affecting adaptation.
Algorithm~\ref{alg:pipeline} summarizes the end-to-end document-only procedure.

\begin{algorithm}[t]
\caption{LLM-based ASR Document-only adaptation.}
\label{alg:pipeline}
\footnotesize
\begin{algorithmic}[1]
\Require source corpus $\dsrc$; target documents $\ctgt=\{d_1,\dots,d_M\}$; pretrained LLM-based ASR; generation LLM $G$ (conversational rendering)
\State \textbf{Render} $\ctgt$ into target adaptation text $\dtgt$, either:
\Statex \quad \emph{Written}: the documents' own sentences; or
\Statex \quad \emph{Conversational}: $N$ dialogs $G(d)$, $d\sim\mathrm{Uniform}(\ctgt)$
\For{each training step}
  \State build batch from $\dsrc$ (views $\sigma_a,\sigma_{ta},\sigma_t$) and
  $\dtgt$ (view $\tau{=}0.5$)
  \State update the LLM-based ASR (denoising)
\EndFor
\State \Return adapted LLM-based ASR
\end{algorithmic}
\end{algorithm}

\section{Building a Document-Grounded Benchmark}
\label{sec:benchmark}
Studying document-only adaptation requires, for each domain, documentation that (a) actually covers the domain's spoken content and (b) is paired with audio and reference transcripts for evaluation.
No public corpus provides this combination.
Crawling the web is also unsuitable here: one of our corpora uses fictitious product and service names that do not exist online, so external documents would not cover the relevant entities.
To obtain a \emph{controlled} instance of such documentation, we synthesize documents from each domain's \emph{training} transcripts only, so no validation or test material enters the pipeline.
To keep the documents from simply echoing the transcripts they replace, generation receives only \emph{abstracted} signals extracted from those transcripts, i.e.\ topic labels, keywords, and entities (steps 2--3 below), and never their raw text or phrasing; the documents thus share with the transcripts only the domain vocabulary, precisely what genuine customer documentation would also contain.
We stress that this construction is used \emph{only to build the benchmark}; the adaptation pipeline of Section~\ref{sec:method} never sees transcripts of the target domain.

The construction proceeds in four steps (Fig.~\ref{fig:pipeline}, top).
(1)~\textbf{Topic discovery:} We cluster the dialogs with a BERTopic-style pipeline~\cite{grootendorst2022bertopic} (class-based TF--IDF), tuning the configuration with Optuna~\cite{akiba2019optuna} ($100$ trials) to maximize $c_v$ coherence over preprocessing, vectorization, dimensionality reduction (LSA, NMF, or UMAP~\cite{mcinnes2018umap}), and clustering (KMeans or HDBSCAN~\cite{mcinnes2017hdbscan}).
(2)~\textbf{Topic labeling:} An LLM labels each topic from its representative dialogs and top keywords.
(3)~\textbf{Entity/keyword extraction:} An LLM extracts, per dialog, the salient domain-specific text spans the ASR must potentially learn, deduplicated into a per-topic set: entities (e.g.\ product, plan, and company names), keywords (e.g.\ domain terms such as ``claim'' or ``premium''), and factual phrases (e.g.\ coverage amounts and waiting periods).
(4)~\textbf{Document generation:} An LLM writes a structured Markdown document per topic from the label, keywords, and entities only (never the raw dialogs), integrating as many as possible and introducing no new facts; documents are capped at $9{,}000$ words and diversified by shuffling the entity list.
After generation, we verify that at least $90\%$ of the extracted entities and keywords appear in the document, regenerating otherwise, so that each document carries enough domain-relevant terminology.
The same \genmodel\ model is used throughout.
We release the generated documents and synthetic dialogs in the SlideSpeech domains to facilitate reproducibility and further research.

\begin{table}[t]
\centering
\caption{Dataset statistics for DefinedAI and SlideSpeech. The base model is trained on the \emph{Source} data and then text-only adapted to each \emph{Target} domain.}
\label{tab:data}
\setlength{\tabcolsep}{2.5pt}
\renewcommand{\arraystretch}{1.3}
\footnotesize
\begin{tabular}{@{}cllrrrrrrr@{}}
\toprule
& \textbf{Role} & \textbf{Domain} & \multicolumn{3}{c}{\textbf{Train}} &
\multicolumn{2}{c}{\textbf{Val}} & \multicolumn{2}{c}{\textbf{Test}}\\
\cmidrule(lr){4-6}\cmidrule(lr){7-8}\cmidrule(lr){9-10}
&  & & \textbf{\#Dlg} & \textbf{\#Utt} & \textbf{Hrs} & \textbf{\#Utt} & \textbf{Hrs} & \textbf{\#Utt} & \textbf{Hrs}\\
\midrule
\multirow{6}{*}{\rotatebox[origin=c]{90}{\textbf{DefinedAI}}}
& Source & B/I/H          & --      & 17{,}398 & 38.2 & -- & --  & -- & --\\
& Target & Banking        & 889     & 26{,}457 & 54.4 & 1{,}443 & 2.9 & 3{,}164 & 6.5\\
& Target & Insurance      & 1{,}126 & 30{,}720 & 65.1 & 1{,}789 & 3.9 & 3{,}667 & 7.7\\
& Target & Healthcare     & 164     & 4{,}702  & 11.5 & 236     & 0.5 & 488     & 1.1\\
& Target & Telco          & 616     & 18{,}835 & 50.8 & 1{,}123 & 3.1 & 2{,}238 & 6.0\\
& Target & Retail         & 354     & 10{,}910 & 27.6 & 618     & 27.6 & 1{,}300 & 3.2\\
\midrule
\multirow{4}{*}{\rotatebox[origin=c]{90}{\textbf{SlideSpeech}}}
& Source & L/T/E          & --  & 34{,}682 & 35.0 & --      & --  & --      & --\\
& Target & Agriculture    & 51  & 30{,}498 & 29.3 & 1{,}679 & 1.7 & 3{,}470 & 3.4\\
& Target & Animation      & 141 & 56{,}593 & 49.8 & 3{,}283 & 2.9 & 7{,}005 & 5.8\\
& Target & Musical Instr. & 20  & 9{,}981  & 8.5  & 852     & 0.7 & 984     & 0.9\\
\bottomrule
\end{tabular}
\end{table}

\section{Experimental Setup}
\label{sec:setup}
\textbf{ASR system and training configuration.} We use the same architecture as the adaptation backbone~\cite{burdisso2025denoising}: the SLAM-ASR framework~\cite{ma2024embarrassingly} with a WavLM-Large encoder~\cite{chen2022wavlm} and a Llama-3.2-3B-Instruct decoder~\cite{dubey2024llama}, connected by a projector of a single hidden layer with a ReLU followed by a regression layer.
We reuse the backbone's hyperparameters unchanged, the only difference being the fixed $\tau=0.5$ (Section~\ref{subsec:backbone}).
The \emph{base model} for each corpus is trained on the source data with the encoder and LLM frozen and only the projector trained, for $5$ epochs with learning rate $10^{-4}$, $1{,}000$ warm-up steps, and batch size $10$.
The text-only denoising \emph{adaptation} fine-tunes the LLM with LoRA on the self-attention query and value projections (rank $16$, $\alpha{=}32$), again for $5$ epochs with learning rate $10^{-4}$ and $1{,}000$ warm-up steps, using the four-view batches of Section~\ref{subsec:backbone}.
As a fully supervised topline we also fine-tune the same LoRA configuration directly on target audio--text pairs (\emph{audio-adapted}); since this uses target audio (the resource our setting aims to replace) and our comparison is among text-only conditions, we report it only on the development domain (Fig.~\ref{fig:nshot}), not in the per-domain tables.

\begin{table}[t]
\centering
\caption{Generated-corpus statistics per target domain.}
\label{tab:gendocs}
\setlength{\tabcolsep}{3.5pt}
\footnotesize
\begin{tabular}{@{}lrrrrrr@{}}
\toprule
\multirow{2}{*}{\textbf{Domain}} & \multirow{2}{*}{\textbf{\#Document}} &
\multicolumn{2}{c}{\textbf{Written}} & \multicolumn{3}{c}{\textbf{Conversational}}\\
\cmidrule(lr){3-4}\cmidrule(lr){5-7}
 & & \textbf{\#Sent} & \textbf{$r_\mathrm{w}$} & \textbf{\#Dlg} & \textbf{\#Utt} & \textbf{$r_\mathrm{c}$}\\
\midrule
Banking        & 23 & 941     & 0.04 & 889     & 43{,}365 & 1.64\\
Insurance      & 25 & 2{,}740 & 0.09 & 1{,}126 & 48{,}540 & 1.58\\
Healthcare     & 14 & 475     & 0.10 & 164     & 7{,}307  & 1.55\\
Telco          & 25 & 5{,}818 & 0.31 & 616     & 27{,}162 & 1.44\\
Retail         & 22 & 1{,}067 & 0.10 & 354     & 12{,}022 & 1.10\\
\midrule
Agriculture    & 21 & 1{,}527 & 0.05 & 51  & 1{,}804 & 0.06\\
Animation      & 25 & 3{,}018 & 0.05 & 141 & 4{,}394 & 0.08\\
Musical Instruments & 19 & 1{,}168 & 0.12 & 20  & 600     & 0.06\\
\bottomrule
\end{tabular}
\end{table}

\textbf{Generation LLM.} Document generation, topic labeling, entity extraction, and document-to-dialog generation all use \genmodel\ via the SDialog toolkit~\cite{sdialog}.

\textbf{Datasets.} We use two corpora (Table~\ref{tab:data}), reporting per domain (\emph{Source} or \emph{Target}) the total number of dialogs (\#Dlg), utterances (\#Utt; the speech segments the ASR is trained and scored on), and hours of audio (Hrs).
\emph{DefinedAI} is a proprietary corpus of manually transcribed customer--agent telephone calls licensed from the data-provider company Defined.ai\footnote{\url{https://defined.ai/}}.
The base model is trained on a \emph{source} mixture of Banking~(B), Insurance~(I), and Healthcare~(H); the \emph{target} domains are these same three (from a disjoint subset, so source and target never overlap) plus the unseen domains Telco and Retail.
\emph{SlideSpeech}~\cite{wang2024slidespeech} is an open, multi-domain corpus of online conference talks and lectures, i.e., presentation-style rather than conversational speech; following~\cite{burdisso2025denoising} we use Life/Talent/English as the source mixture and Agriculture (Ag), Animation (An), and Musical Instruments (MI) as target domains.
The two corpora thus differ acoustically and stylistically (telephone customer-service calls vs.\ presentation-style talks), so the cross-domain setting (Section~\ref{sec:results}) compounds lexical and recording-condition shifts.

\textbf{Adaptation conditions (text source).} Table~\ref{tab:gendocs} reports what the two generation pipelines produced as a corpus for domain adaptation: the number of documents (\#Document) and their sentences (\#Sent), with \#Dlg and \#Utt as in Table~\ref{tab:data}; $r_\mathrm{w}$ and $r_\mathrm{c}$ give \#Sent and \#Utt as fractions of the real training \#Utt from Table~\ref{tab:data}.
The generated documents are compact ($14$--$25$ per domain): the \#Sent pool is well below the real \#Utt ($r_\mathrm{w}\le0.31$), while the \#Utt pool is comparable to real on DefinedAI ($r_\mathrm{c}\approx1.1$--$1.6$) but far smaller on SlideSpeech ($r_\mathrm{c}\approx0.06$).
We compare four experimental conditions, all using the \emph{same} denoising backbone and $\tau=0.5$: (i)~\emph{Base}, the unadapted source model; (ii)~\emph{Transcripts}, adaptation on ground-truth target transcriptions (\#Utt, Table~\ref{tab:data}), the upper bound for text-only adaptation; (iii)~\emph{Written}, the documents split into sentences and used directly as the target text (\#Sent, Table~\ref{tab:gendocs}); (iv)~\emph{Conversational}, the same documents rendered as synthetic dialogs, using their utterances as the target text (\#Utt, Table~\ref{tab:gendocs}).
Conditions (iii) and (iv) are the \emph{document-only} settings (no transcripts); (iii) is included to test the natural alternative of fine-tuning on the document text itself.


\textbf{Metric.} We report word error rate (WER, \%) and, where useful, the relative improvement $\Delta$ over the corresponding base model.

\textbf{Reproducibility.} The generation, clustering, labeling, extraction, and document-creation pipelines are implemented on top of the SDialog toolkit~\cite{sdialog} with fixed random seeds.
We release all code (generation, benchmark construction, and experiments) together with the generated synthetic dialogs as well as the SlideSpeech documents for all its domains.

\section{Results}
\label{sec:results}
All result tables use the four conditions from the previous section: \emph{Base}, \emph{Transcripts}, and the \emph{Document-only} pair \emph{Written} and \emph{Conversational}; $\Delta$ is the relative WER improvement over \emph{Base}; and \textbf{bold} marks the better document-only rendering, including near-ties within $0.1$ WER.
We organize results by corpus: adaptation within DefinedAI, whose targets include domains seen in the source (Banking, Insurance, Healthcare) as well as unseen ones (Telco, Retail, marked \emph{ood}), and within SlideSpeech (all targets unseen), followed by the harder cross-domain transfer between the two corpora.

\subsection{DefinedAI}
Table~\ref{tab:defai} reports WER on the DefinedAI targets, where the base model is trained on the DefinedAI source.
Two findings stand out.
First, the conversational condition recovers most of the gain of real transcriptions: across the five domains the conversational--transcript WER gap is at most $0.28$ absolute (Banking), and on Retail it even surpasses the transcript condition ($17.18$ vs.\ $17.59$).
Second, the written condition is consistently far weaker: near or below the base model on Banking and Insurance, and \emph{worse} than the base model on Healthcare ($7.63$ vs.\ $6.85$).
The documents carry the relevant entities and keywords by construction (Sec.~\ref{sec:benchmark}), but their written form lacks the spoken characteristics the ASR must model; the conversational view of the same content is far more effective here.

\begin{table}[t]
\centering
\caption{WER (\%) on DefinedAI; $\Delta$: relative improvement over base. Telco
and Retail are unseen domains (ood).}
\label{tab:defai}
\setlength{\tabcolsep}{4pt}
\footnotesize
\begin{tabular}{@{}lrrr|rrrr@{}}
\toprule
\multirow{3}{*}{\textbf{Domain}} & \multirow{3}{*}{\textbf{Base}} &
\multicolumn{2}{c|}{} & \multicolumn{4}{c}{\textbf{Document-only}}\\
\cmidrule(l){5-8}
 & & \multicolumn{2}{c|}{\textbf{Transcripts}} &
\multicolumn{2}{c}{\textbf{Written}} & \multicolumn{2}{c}{\textbf{Conversational}}\\
\cmidrule(lr){3-4}\cmidrule(lr){5-6}\cmidrule(lr){7-8}
 & & WER & $\Delta$ & WER & $\Delta$ & WER & $\Delta$\\
\midrule
Banking      & 10.90 & 9.13 & 16.2 & 10.93 & $-0.3$ & \textbf{9.41} & 13.7\\
Insurance    & 10.15 & 8.84 & 12.9 & 9.87  & 2.8   & \textbf{8.89} & 12.4\\
Healthcare   & 6.85  & 6.13 & 10.5 & 7.63  & $-11.4$ & \textbf{6.29} & 8.2\\
Telco (ood)  & 20.12 & 17.61& 12.5 & 19.19 & 4.6   & \textbf{17.86} & 11.2\\
Retail (ood) & 20.48 & 17.59& 14.1 & 18.79 & 8.3   & \textbf{17.18} & 16.1\\
\bottomrule
\end{tabular}
\end{table}

\subsection{SlideSpeech}
Table~\ref{tab:ss} reports WER on the SlideSpeech targets (all unseen), with the base trained on the SlideSpeech source.
Here even real transcripts gain little ($\Delta\le7.1\%$), so the headroom for any text-only method is small; within it, two effects stand out.
First, unlike on DefinedAI, the written condition is competitive: SlideSpeech consists of presentation-style talks and lectures~\cite{wang2024slidespeech} rather than two-party conversations, so the documents' expository sentences already match the presentation register, whereas the customer-service dialog form produced by our DefinedAI-tuned prompt is a weaker fit.
Second, the default conversational setting is data-limited: SlideSpeech has few dialogs, so generating $N$ equal to the real dialog count yields only $51$--$141$ dialogs and a few thousand utterances (Table~\ref{tab:gendocs}), far below the tens of thousands in the real data.
Because generation is decoupled from the document count, we can simply generate more: raising Agriculture and Animation to $1{,}000$ dialogs ($40$k and $31$k utterances, matching the real data's order of magnitude) lowers WER from $16.27$/$16.13$ to $16.17$/$15.68$, approaching real transcriptions ($16.01$/$15.55$), with the larger gain on Animation.
When headroom is small, then, the \emph{amount} of synthetic data, not only its form, governs how much is recovered, and the conversational view offers a tunable knob to supply it.

\begin{table}[t]
\centering
\caption{WER (\%) on SlideSpeech; all targets are unseen (ood).}
\label{tab:ss}
\setlength{\tabcolsep}{4pt}
\footnotesize
\begin{tabular}{@{}lrrr|rrrr@{}}
\toprule
\multirow{3}{*}{\textbf{Domain}} & \multirow{3}{*}{\textbf{Base}} &
\multicolumn{2}{c|}{} & \multicolumn{4}{c}{\textbf{Document-only}}\\
\cmidrule(l){5-8}
 & & \multicolumn{2}{c|}{\textbf{Transcripts}} &
\multicolumn{2}{c}{\textbf{Written}} & \multicolumn{2}{c}{\textbf{Conversational}}\\
\cmidrule(lr){3-4}\cmidrule(lr){5-6}\cmidrule(lr){7-8}
 & & WER & $\Delta$ & WER & $\Delta$ & WER & $\Delta$\\
\midrule
Agriculture        & 16.25 & 16.01 & 1.5 & \textbf{16.09} & 1.0 & 16.27 & $-0.1$\\
Animation          & 16.45 & 15.55 & 5.5 & \textbf{16.22} & 1.4 & \textbf{16.13} & 1.9\\
Musical Instr.\    & 15.16 & 14.09 & 7.1 & 14.84 & 2.1 & \textbf{14.15 }& 6.7\\
\bottomrule
\end{tabular}
\end{table}

\subsection{Cross-domain (DefinedAI\,$\rightarrow$\,SlideSpeech)}
The hardest setting adapts a DefinedAI-source model to SlideSpeech targets, combining a domain shift with an acoustic/recording-condition shift, so the base WER is very high ($47$--$59$).
Table~\ref{tab:cross} shows all three text sources roughly halve the WER and land close together: over the base, the conversational condition improves by $44.9$/$43.2$/$39.9\%$ and the written condition by $44.1$/$38.0$/$45.3\%$, versus $49.5$/$39.6$/$45.2\%$ for real transcriptions; the conversational condition surpasses real on Animation ($27.36$ vs.\ $29.12$), while written and real perform virtually the same on Musical Instruments ($25.73$ vs.\ $25.76$).
Strikingly, this holds with the \emph{default} small dialog count: because the source-trained base is far from the targets, even a little in-domain text recovers most of a large gain, in contrast to the data-limited in-corpus setting above.
This is our strongest evidence: starting from a model trained on an unrelated corpus and using \emph{only} target documents, one recovers most of the improvement that would otherwise require real in-domain transcripts, exactly the onboarding scenario our work targets.

\begin{table}[t]
\centering
\caption{Cross-domain WER (\%): base trained on DefinedAI source, adapted to
SlideSpeech targets.}
\label{tab:cross}
\setlength{\tabcolsep}{4pt}
\footnotesize
\begin{tabular}{@{}lrrr|rrrr@{}}
\toprule
\multirow{3}{*}{\textbf{Domain}} & \multirow{3}{*}{\textbf{Base}} &
\multicolumn{2}{c|}{} & \multicolumn{4}{c}{\textbf{Document-only}}\\
\cmidrule(l){5-8}
 & & \multicolumn{2}{c|}{\textbf{Transcripts}} &
\multicolumn{2}{c}{\textbf{Written}} & \multicolumn{2}{c}{\textbf{Conversational}}\\
\cmidrule(lr){3-4}\cmidrule(lr){5-6}\cmidrule(lr){7-8}
 & & WER & $\Delta$ & WER & $\Delta$ & WER & $\Delta$\\
\midrule
Agriculture        & 59.40 & 29.98 & 49.5 & 33.22 & 44.1 & \textbf{32.70} & 44.9\\
Animation          & 48.18 & 29.12 & 39.6 & 29.86 & 38.0 & \textbf{27.36} & 43.2\\
Musical Instr.\    & 47.02 & 25.76 & 45.2 & \textbf{25.73} & 45.3 & 28.25 & 39.9\\
\bottomrule
\end{tabular}
\end{table}

\subsection{Effect of few-shot prompting}
\label{subsec:nshot}
\begin{figure}[t]
\centering
\begin{tikzpicture}
\begin{axis}[
  ybar, bar width=8pt,
  width=\columnwidth, height=4.5cm,
  enlarge x limits=0.12,
  symbolic x coords={Audio,Transcripts,0-shot,1-shot,3-shot,10-shot},
  xtick={Audio,Transcripts,0-shot,1-shot,3-shot,10-shot},
  x tick label style={font=\scriptsize,rotate=25,anchor=north east,yshift=1pt},
  ylabel={Average WER (\%)},
  ymin=8.4, ymax=11,
  ytick={8.5,9,9.5,10,10.5,11},
  tick label style={font=\scriptsize}, label style={font=\scriptsize},
  nodes near coords, every node near coord/.append style=
    {font=\tiny,anchor=south,yshift=1pt},
  error bars/y dir=both, error bars/y explicit,
  grid=both, grid style={black!8},
]
\addplot[ybar, bar shift=0pt, fill=black!18, draw=black!40] coordinates {
  (Audio,8.68)       +- (0,0.03)
  (Transcripts,9.13) +- (0,0.04)
};
\addplot[ybar, bar shift=0pt, fill=violet!45, draw=violet!60] coordinates {
  (0-shot,9.41)      +- (0,0.04)
  (1-shot,9.23)      +- (0,0.11)
  (3-shot,9.21)      +- (0,0.16)
  (10-shot,9.42)     +- (0,0.27)
};
\draw[dashed,thick,gray!80!black]
  ({rel axis cs:0,0} |- {axis cs:Audio,10.9}) --
  ({rel axis cs:1,0} |- {axis cs:Audio,10.9});
\node[anchor=north east,font=\tiny,text=gray!80!black]
  at (axis cs:10-shot,10.9) {Base = 10.90};
\end{axis}
\end{tikzpicture}
\caption{Banking (development domain) average WER by adaptation text source;
all conditions are averaged over three seeds (error bars).
Light gray bars (\emph{Audio}, audio-adapted; \emph{Transcripts}, real transcriptions) are reference toplines our method avoids.
Colored bars are our \emph{Conversational} document-only condition with $0$/$1$/$3$/$10$ real example dialogs in the prompt ($0$-shot, the transcript-free setting used throughout).
The dashed line marks the unadapted base model.}
\label{fig:nshot}
\end{figure}
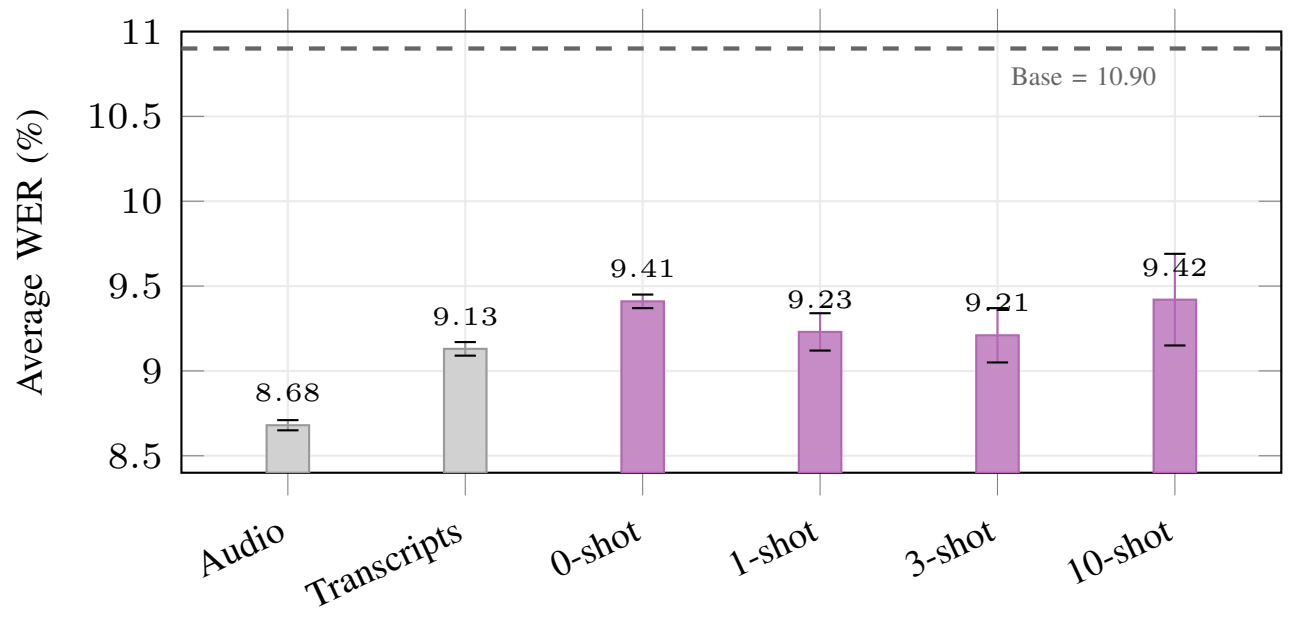

We also test whether seeding the document-to-dialog prompt with a few real example dialogs ($n$-shot) helps, on Banking (Fig.~\ref{fig:nshot}).
All conditions use three seeds; few-shot adds a second source of randomness, since each seed samples different example dialogs.
Few-shot changes WER only marginally ($1$/$3$-shot $9.23$/ $9.21$ vs.\ zero-shot $9.41$; $10$-shot $9.42$), with all conditions within $0.3$ WER of the \emph{Transcripts} topline ($9.13$) and the best ($3$-shot) within $0.1$; $10$-shot adds no gain and is the most variable, plausibly because its longer prompt (ten example dialogs) is harder to use reliably.
Since \emph{any} few-shot variant requires real in-domain transcripts (violating our document-only premise) and the gap is small, we adopt zero-shot generation throughout.
 
\subsection{Prompt design}
\label{subsec:prompt}
We stress that our goal is \emph{feasibility}, not prompt optimality.
We designed the generation prompt of Section~\ref{subsec:doc2dialog} once, on Banking (our development domain), guided by lightweight checks that the synthetic dialogs matched real ones (turn-length, $n$-gram perplexity, and authorship style~\cite{rivera2021luar}) together with downstream Banking WER, then \emph{froze} it across all domains and corpora.
Since it was never tuned per domain and is likely suboptimal, the reported results are a conservative lower bound; refining the prompt is an orthogonal direction left to future work.

\subsection{Discussion}

Across all three settings (Tables~\ref{tab:defai}--\ref{tab:cross}), document-only adaptation recovers much of the improvement real transcripts give, but which rendering wins, and how much text it needs, depends on the target through three factors.
\textbf{Content} is fixed by construction: every condition carries the same entities and keywords.
\textbf{Form} is decisive for conversational speech: on DefinedAI the written sentences barely help and on Healthcare even hurt, whereas rendering the same content as dialogs recovers nearly all the transcript gain. 
On the presentation-style SlideSpeech targets form matters far less, since the documents' expository sentences already fit that register and the written condition stays competitive (Table~\ref{tab:ss}).
\textbf{Quantity} takes over when the form fits but the pool is small: it is roughly real-sized on DefinedAI ($r_\mathrm{c}\approx1$) but tiny on SlideSpeech ($\approx0.06$, Table~\ref{tab:gendocs}), where scaling generation to $1{,}000$ dialogs improves the two domains with room to spare, most clearly Animation ($16.13\rightarrow15.68$, near its $15.55$ transcript WER, Table~\ref{tab:ss}). When the base instead starts far from the target (cross-domain), the headroom is so large that even the small default recovers most of it, with all text sources yielding similar gain over base (Table~\ref{tab:cross}).
In short, the conversational view helps in two ways at once, turning compact documentation into spoken-style text and supplying as much of it as needed.
These results hold with a single, fixed, non-tuned prompt (Section~\ref{subsec:prompt}), suggesting that document-only adaptation is robust rather than an artifact of prompt engineering.

\section{Conclusion}
\label{sec:conclusion}
We introduced document-only adaptation of LLM-based ASR: adapting to a new domain from raw documents alone, with no target audio and no target transcripts.
We use a domain's documents as adaptation text in two ways: directly, as their own written sentences, or rendered into synthetic spoken dialogs.
Fed to a denoising-based adaptation method, the conversational rendering generally approaches, and sometimes matches or exceeds, ground-truth-transcript adaptation across the three settings on two corpora; the written rendering is competitive when its register fits the target or the model starts far from it, but can fall short, or even hurt, otherwise.
This makes adaptation feasible from documentation a customer already owns, without recording or transcribing sensitive conversations.
We release the generated SlideSpeech documents and dialogs to support future research.

\section*{Acknowledgment}

This work was supported by Idiap Research Institute and Uniphore collaboration project. Part of this work was also supported by EU Horizon 2020 project ELOQUENCE (grant number 101070558).
During the preparation of this manuscript, the authors used Claude Opus 4.8 (Anthropic) to assist with editing, rephrasing, and polishing the writing throughout the paper. The authors reviewed and edited all AI-assisted text and take full responsibility for the content of this article.

\bibliographystyle{IEEEtran}
\bibliography{refs}

\end{document}